\pdfoutput=1
\documentclass[runningheads,a4paper]{llncs}
\usepackage{amssymb}
\usepackage{times}
\usepackage{graphicx}
\usepackage{subcaption}
\usepackage{booktabs}
\usepackage[numbers]{natbib}
\usepackage{algorithm}
\usepackage{algorithmic}
\usepackage{microtype}
\usepackage{todonotes}
\usepackage{siunitx}

\tikzset{
	declare function={
		mysign(\x) = (and(\x<=0, 1) * -1) +
		       (and(\x>0, 1) * 1);
		}
	}

\usepackage{hyperref}

\usepackage{glossaries}
\usepackage{algorithm}
\usepackage{multirow}
\usepackage{footnote}
\usepackage{rotating}
\makesavenoteenv{tabular}
\makesavenoteenv{table}
\newcommand{\dblquotes}[1]{``#1''}
\newcommand{\OurScheme}{\textsc{Finn}}
\newcommand{\FINN}{\OurScheme{}}

\usepackage{url}
\urldef{\mailsa}\path|j.su13@ic.ac.uk|    
\urldef{\mailsb}\path|{nfraser,giulio,mblott}@xilinx.com|

\newcommand{\keywords}[1]{\par\addvspace\baselineskip
\noindent\keywordname\enspace\ignorespaces#1}

\begin{document}

\mainmatter  

\title{Accuracy to Throughput Trade-offs for Reduced Precision Neural Networks on Reconfigurable Logic}

\titlerunning{Accuracy to Throughput Trade-offs for Reduced Precision NN on FPGA}

%
%
\author{Jiang Su \and Nicholas J. Fraser \and Giulio Gambardella \and Michaela Blott \and Gianluca Durelli \and David B. Thomas \and Philip Leong \and Peter Y. K. Cheung}

\authorrunning{J. Su, N. Fraser, G. Gambardella, M. Blott, \textit{et. al.}}

\institute{Xilinx Research Labs, Imperial College London, University of Sydney \\
\mailsa\\
\mailsb\\
}

%
%

\maketitle

\glsdisablehyper
\newacronym{FPGA}{FPGA}{Field Programmable Gate Array}
\newacronym{CNN}{CNN}{Convolutional Neural Network}
\newacronym{BNN}{BNN}{Binarized Neural Network}
\newacronym{OFM}{OFM}{Output Feature Map}
\newacronym{IFM}{IFM}{Input Feature Map}
\newacronym{OCM}{OCM}{On-Chip Memory}
\newacronym{MVU}{MVTU}{Matrix--Vector--Threshold Unit}
\newacronym{SWU}{SWU}{Sliding Window Unit}
\newacronym{TU}{TU}{Thresholding Unit}
\newacronym{PU}{PU}{Pooling Unit}
\newacronym{II}{II}{initiation interval}
\newacronym{PE}{PE}{Processing Element}
\newacronym{NN}{NN}{Neural Network}
\newacronym{ANN}{ANN}{Artificial Neural Network}
\newacronym{FPS}{FPS}{frames per second}
\newacronym{HLS}{HLS}{High-Level Synthesis}
\newacronym{ILSVRC}{ILSVRC}{ImageNet Large Scale Visual Recognition Competition}
\newacronym{TOPS}{TOPS}{teraoperations per second}
\newacronym{GOP}{GOPS}{billion operations}
\newacronym{GFLOP}{GFLOP}{billion floating point operations}
\newacronym{LUTs}{LUT}{look up table}
\newacronym{MAC}{MAC}{multiply--accumulate}
\newacronym{RELU}{ReLU}{Rectified Linear Unit}
\newacronym{HARDTANH}{hard-tanh}{Hard Hyperbolic Tangent Function}
\begin{abstract}
Modern \glspl{CNN} are typically based on floating point linear algebra based implementations. Recently, reduced precision \glspl{NN} have been gaining popularity as they require significantly less memory and computational resources compared to floating point. This is particularly important in power constrained compute environments. However, in many cases a reduction in precision comes at a small cost to the accuracy of the resultant network. In this work, we investigate the accuracy-throughput trade-off for various parameter precision applied to different types of NN models. We firstly propose a quantization training strategy that allows reduced precision NN inference with a lower memory footprint and competitive model accuracy. Then, we quantitatively formulate the relationship between data representation and hardware efficiency. Our experiments finally provide insightful observation. For example, one of our tests show 32-bit floating point is more hardware efficient than 1-bit parameters to achieve 99\% MNIST accuracy. In general, 2-bit and 4-bit fixed point parameters show better hardware trade-off on small-scale datasets like MNIST and CIFAR-10 while 4-bit provide the best trade-off in large-scale tasks like AlexNet on ImageNet dataset within our tested problem domain.

\keywords{Reduced precision, neural networks, FPGA, algorithm acceleration}
\end{abstract} 
\section{Introduction}
\label{intro}

Modern \glspl{CNN} may contain millions of floating-point parameters and require billions of floating-point operations to recognize a single image. These requirements tend to increase as researchers explore deeper networks. On the other hand, the integration of computing resources on hardware platforms is hampered by the slowing down of Moore’s law. Therefore, it is meaningful to study efficient model designs with customized data paths and effective data representations.

\begin{figure}
\begin{centering}
\includegraphics[height=4cm]{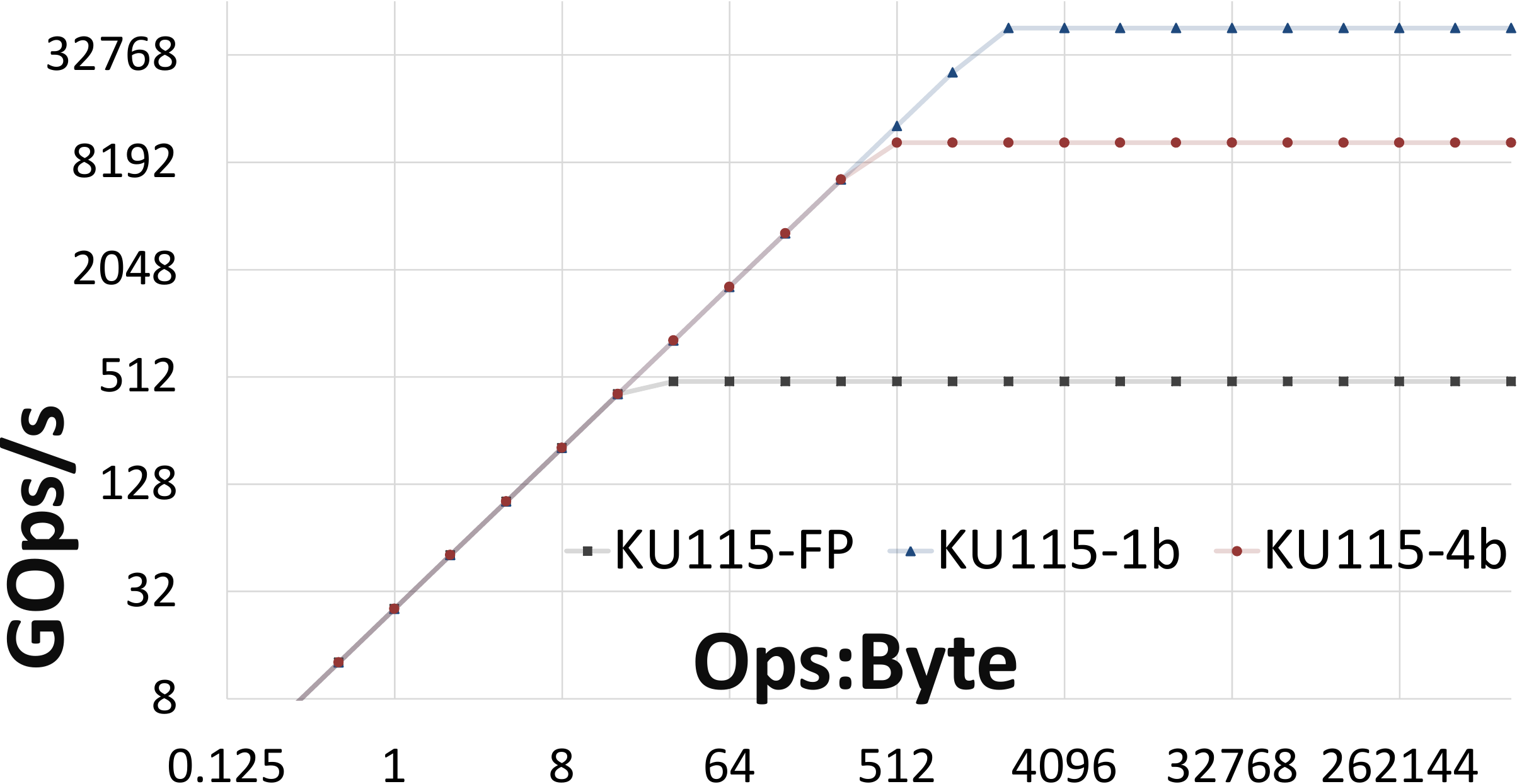}\\
\caption{Roofline Model for Xilinx KU115 FPGA}
\vspace{-20pt}
\label{roofline}
\end{centering}

\end{figure}

Previous work showed that using reduced precision for NN parameters provide massive improvements on system performance such as throughput, computational resource usage and memory footprint \cite{finn,binarynet,xnornet}. For example, Figure \ref{roofline} shows the roofline for Xilinx KU115 device in terms of its arithmetic intensity and peak board performance. It shows that higher performance ``ceiling" can be achieved if using lower precision data in operations. However, as mentioned in \cite{ResiliencyUnderQuantization} and \cite{finn}, reduced-precision parameters need more operations and parameters to achieve the same accuracy provided by high precision alternatives. Additionally, the ``operation" as for the y axis in Figure \ref{roofline} can be different to various data types. For example, instead of expensive Multiply Accumulate (MAC) operations for floating point (FP) or fixed point (FIX) representations, XNOR and popcount logic can be used for Binary Neural Networks (BNNs). Therefore,  compared to FP, binary number operations may lead to a system with higher throughput (GOps/s), but does this higher throughput provide as good NN accuracy? Another way to ask the question is that if a target classification accuracy is given to a particular dataset, can binary parameter based NN still allow more efficient hardware systems than floating point parameters? We found many questions like this remain unanswered. For example, how does parameter precision in NNs affect the hardware throughput given a particular system architecture? Which data type provides the best trade-off between model accuracy and hardware performance?

In order to address above questions, we focus on an exploration space for various data representations in NN computation in order to study their impacts to hardware system efficiency and model accuracy. In contrast to previously published work, which focuses either on hardware-wise efficiency \cite{dorefa}\cite{finn}\cite{binarynet} or model performance \cite{denil1014}\cite{hwang1014}\cite{wu0516}, we consider both perspectives and tentatively provide a more comprehensive view of using reduced-precision for NN system design. The contributions of this work is as follows:

\begin{itemize}
    \item We report our quantization training strategy for NN inference with quantized weights and activations in arbitrary precision types. Without any compression techniques, our training strategy requires less memory footprint and achieves competitive accuracy compared to several state-of-the-art compression techniques on the same task.
    \item We propose quantitative estimation models to show how parameter precision affects hardware cost and system throughput for a NN hardware system.
    \item We publish systematic experimental results for different types of NNs with weights and activations represented separately in 1-bit (Binary), 2-bit (INT2), 4-bit (INT4), 8-bit(INT8), 16-bit (INT16) fixed point values and 32-bit floating point values (FP32) and show their impacts to classification accuracy, hardware cost and inference throughput.
    \item Finally, our exploration space provides useful insights and a more comprehensive view of using reduced-precision values in NN acceleration. For example, in our MNIST experiments, a networks with FP32 parameters is more memory efficient than 1-bit parameters for achieving ~99\% accuracy due to the smaller topology required. In general, 2-bit and 4-bit fixed point parameters show better hardware trade-off on small-scale datasets like MNIST and CIFAR-10 while 4-bit provide the best trade-off in large-scale tasks like AlexNet on ImageNet dataset within our tested problem domain.
\end{itemize}

In the next section, we introduce our training strategy for reduced precision parameters. Next, Section \ref{hw_arch} introduces the proposed estimation models for hardware cost and system throughput. The experimental results are discussed in Section \ref{experiments} and Section \ref{conclusion} finally concludes the paper.
\section{Training Strategies}
\label{methodology}

In this work, weights and activation values are quantized before used in the feedforward and backward propagation. For fixed-point representations, values are represented with $WL$ bits, in which the Most Significant Bit (MSB) indicates the sign while $FL$ and $(WL-FL-1)$ bits are used for expressing the fractional and the integer parts separately. 

Specifically, for binary representation, we adopted the deterministic binarization function used in \cite{binarynet} as our quantization method:

\begin{equation}
    x^Q = Sign(x) = \begin{cases}
+1 & \text{ if } x\geq 0 \\ 
-1 & \text{ otherwise. }  
\end{cases}
\label{quant_bin}
\end{equation}

For fixed point values, the quantization function converts real values to nearest pre-defined fixed point representations. 

As mentioned in \cite{binarynet}\cite{finn}, for training binary parameters, batch normalization is generally conducted before the activation function while for other representations, it's the other way around. For activation functions, we use both of the \gls{HARDTANH} $\sigma(x)=Min(1, Max(-1,x))$ and the \gls{RELU} $\sigma(x) = Max(0,x)$. Both of the activations are used in our experiments and the one that delivers a higher model accuracy is selected.

\begin{algorithm}[ht]
\caption{Quantization training strategy for an $L$-layer neural network }
\begin{algorithmic}[1]
\label{training}
\REQUIRE At time step $t$, a batch of inputs $a_0$ and their labels $a^*$, network weights $W$, Batch Normalization parameter $\theta$, learning rate $\eta$ and its decay factor $\lambda$.
\ENSURE At time step $t+1$, the updated weights $W^{t+1}$, the updated Batch Normalization parameter $\theta^{t+1}$ and the updated learning rate $\eta^{t+1}$.

\COMMENT{1. Propagations with Limited Precision Parameters}

\COMMENT{1.a Feedforward Propagation:}

\FOR{$i=1$\text{ to }$L$}
\STATE $W_i^Q\gets Quantize(W_i)$
\STATE $s_i\gets a_{i-1}^Q*W_i^Q$
\IF{$i<L$}
\STATE $a_i\gets ActFunc(\hat{s}_i)$
\STATE $\hat{s}_i\gets BatchNorm(s_i,\theta_i)$
\STATE $a_i^Q \gets Quantize(a_i)$
\ENDIF
\ENDFOR

\COMMENT{1.b Backward Propagation:}

Compute $g_{a_L}=\frac{\partial C}{\partial a_L}$ knowing $a_L$ and $a^*$
\FOR{$i=L$\text{ to }$1$}
\IF{$i<L$}
\STATE $(g_{s_i},g_{\theta_i})\gets BackBatchNorm(\hat{g}_{a_i},s_i,\theta_i)$
\STATE $\hat{g}_{a_i}\gets BackActFunc(g_{a_i})$
\ENDIF
\STATE $g_{a_{i-1}^Q}\gets g_{s_i}*W_i^Q$
\STATE $g_{W_i^Q}\gets g_{s_i}^T*a_{i-1}^Q$
\ENDFOR

\COMMENT{2. Weight Updating with High Precision Parameters}

\FOR{$i=1$\text{ to }$L$}
\STATE $\theta_i^{t+1}\gets Update(\theta_i,\eta,g_{\theta_i})$
\STATE $W_i^{t+1}\gets Clip(Update(W_i,\gamma_i\eta,g_{W_i^Q}),-1,1)$
\STATE $\eta^{t+1}\gets \lambda \eta$
\ENDFOR
\end{algorithmic}
\end{algorithm}

Globally, quantized low-precision weights and activations are used for feedforward and backpropagation passes. After this process, the floating point parameters are updated accordingly (line 20, Algorithm 1). High-precision values are used for updating because they can accumulate tiny value changes while lower-precision values can improve the computational efficiency during inference due to the low design complexity \cite{courbariaux0915}.

Our quantization training process is shown in Algorithm \ref{training}. $Quantize(*)$ is the quantization function. $BatchNorm(*)$ and $BackBatchNorm(*)$ are functions that propagate neuron-generated values and gradients separately in feedforward and backpropagations. Similarly, $ActFunc(*)$ and $BackActFunc(*)$ are activation passes in above-mentioned bidirectional propagations. $Update(*)$ specifies the parameter updating strategy, ADAM Updating is used in this work\cite{adam}. Network weights are initialized based on \cite{Glorot10}. Finally, $C$ is the cost function.

\begin{table}
    \centering
    \caption{The Expected Cost Per Operation For Each Precision Type}
\begin{tabular}{lrrrrrrrrr}
\toprule
Datatype & LUTs & LUTs & LUTs & DSPs  & DSPs  & DSPs  & $C_{avg}$       & $C_{rel}$\\
         & min  & max  & avg  & min   & max   & avg   & $\times 10^{-6}$&\\
\midrule
Binary   & 4.24 & 8.00 & 5.58 & 0     & 0     & 0     & 12.02         & 1\\
INT2     & 10.98 & 18.74 & 13.52 & 0     & 0     & 0     & 29.12        & 2.42\\
INT4     & 27.18 & 35.56 & 30.06& 0     & 0     & 0     & 64.76        & 5.39\\
INT8     & 83.28& 91.92& 86.38& 0     & 0     & 0     & 186.02        & 15.48\\
INT16    & 21.64& 38.36& 28.66& 1   &1   & 1   & 181.16        & 15.07\\
FP32  & 356  & -    & -    & 4     & -     & -     & 766.6        & 63.79\\
\midrule
KU115& -    & 663,360 & - & -     & 5,520 & -     & -            & -\\
\bottomrule
\end{tabular}

    \label{tab:op-cost}
    \vspace{-20pt}
\end{table}

In the feedforward process, real-valued weights $W$ are firstly quantized into low-precision weights $W^Q$ as shown in line 2. After batch normalization and activation function, neuron activations are also quantized to low precision (line 7). Above steps form a layer-wise process until the training error $g_{a_L}$ is calculated in the last layer according to the outputs in the output layer $a_L$ and the corresponding data label $a*$. Then backward propagation starts with the error calculated through above feedforward pass. After going through backward passes for the activation function and batch normalization function, the quantized weights $W^Q$ are used for the calculation of gradients of both neurons and connections. Noticeably, this is the key point that the model is ``aware" of the quantized parameters. This process is a layer-wise process from the output layer to input layer (line 10 - 17). Finally, parameters are updated with the gradients following the ADAM rule. Specifically, the updated values are clipped between -1 and +1 for regularization. The ADAM parameter $\theta$ and learning rate $\eta$ are also updated accordingly. If the $ActFunc(*)$ is hard-tanh and $Quantize(*)$ is Eq.\ref{quant_bin}, the algorithm \ref{training} depicts the training strategy proposed in \cite{binarynet} for 1 bit binary parameters.

The quantization training process is done offline and we only deploy the inference process online with the trained parameters. In the next section, a hardware cost model is introduced for a specific system architecture on FPGAs.
\section{Hardware Cost Model For Different Precision Types}
\label{hw_arch}

In this work, we build up our hardware impact analysis based on a hardware system architecture that is introduced in this section. Firstly, processing elements are introduced as the basic building blocks of conducting above operations. Then a hardware cost model is proposed to theoretically formulate the relationship between parameter precision and system throughput, which is then later applied to our studied trade-offs.

\subsection{System Architecture}
\label{system}

\begin{figure*}
	\centering
	\begin{subfigure}[b]{0.45\linewidth}
		\centering
		\includegraphics[height=2.4cm]{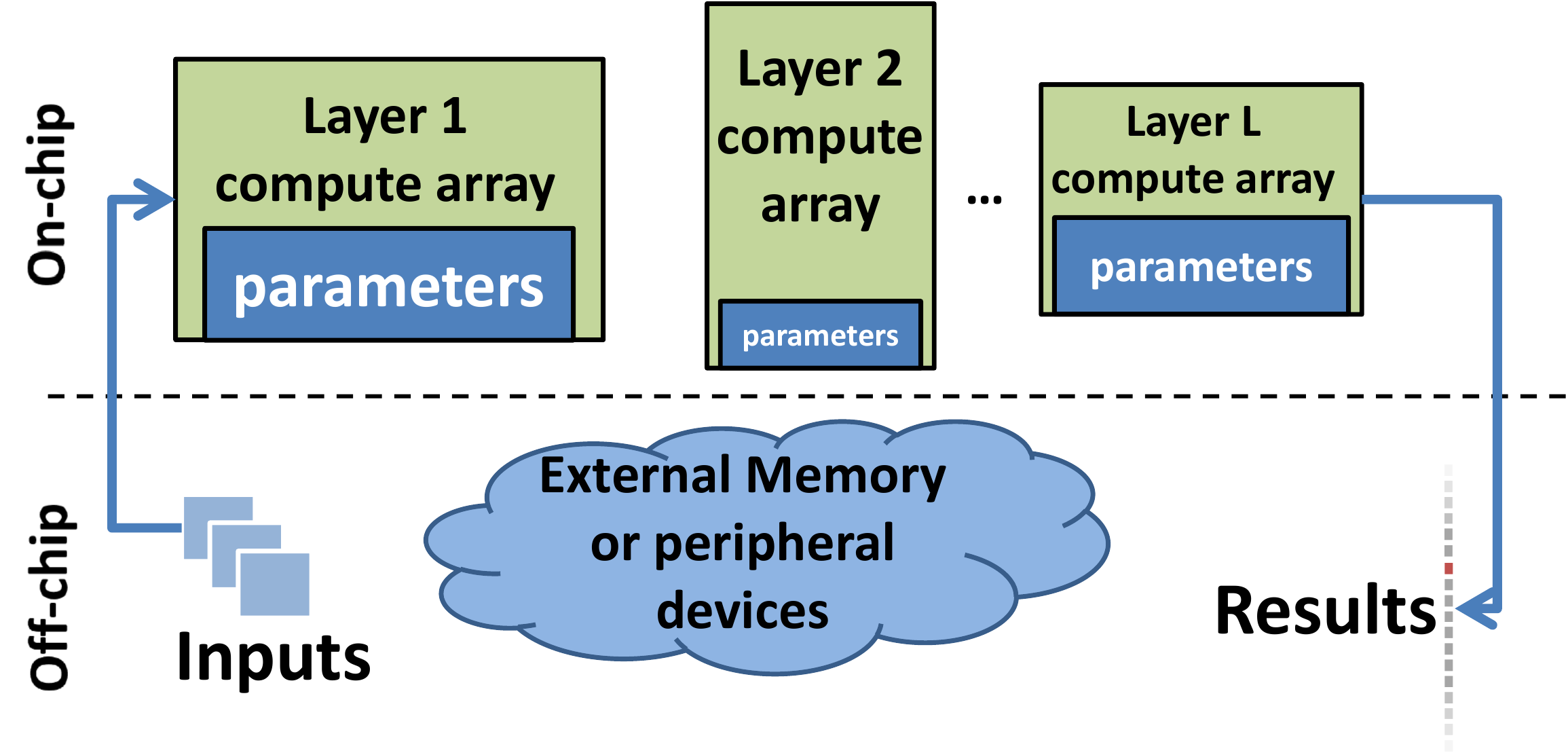}
		\caption{Top-Level Architecture.}
		\label{fig:arch}
	\end{subfigure}
	\begin{subfigure}[b]{0.45\linewidth}
		\centering
		\includegraphics[height=2.58cm]{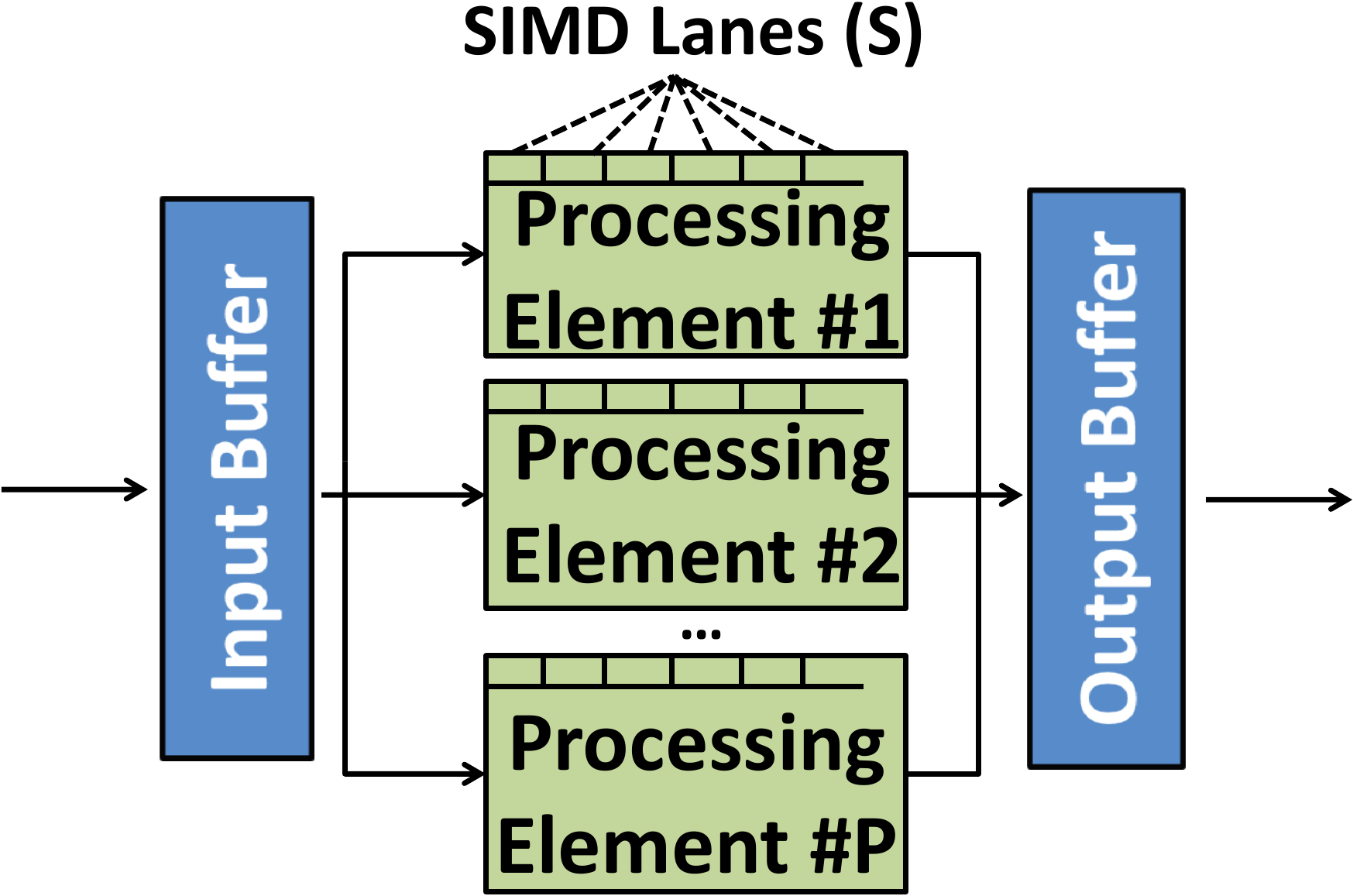}
		\caption{MVOU}
		\label{fig:mvtu}
	\end{subfigure}
	\begin{subfigure}[b]{0.45\linewidth}
		\centering
		\includegraphics[height=2.5cm]{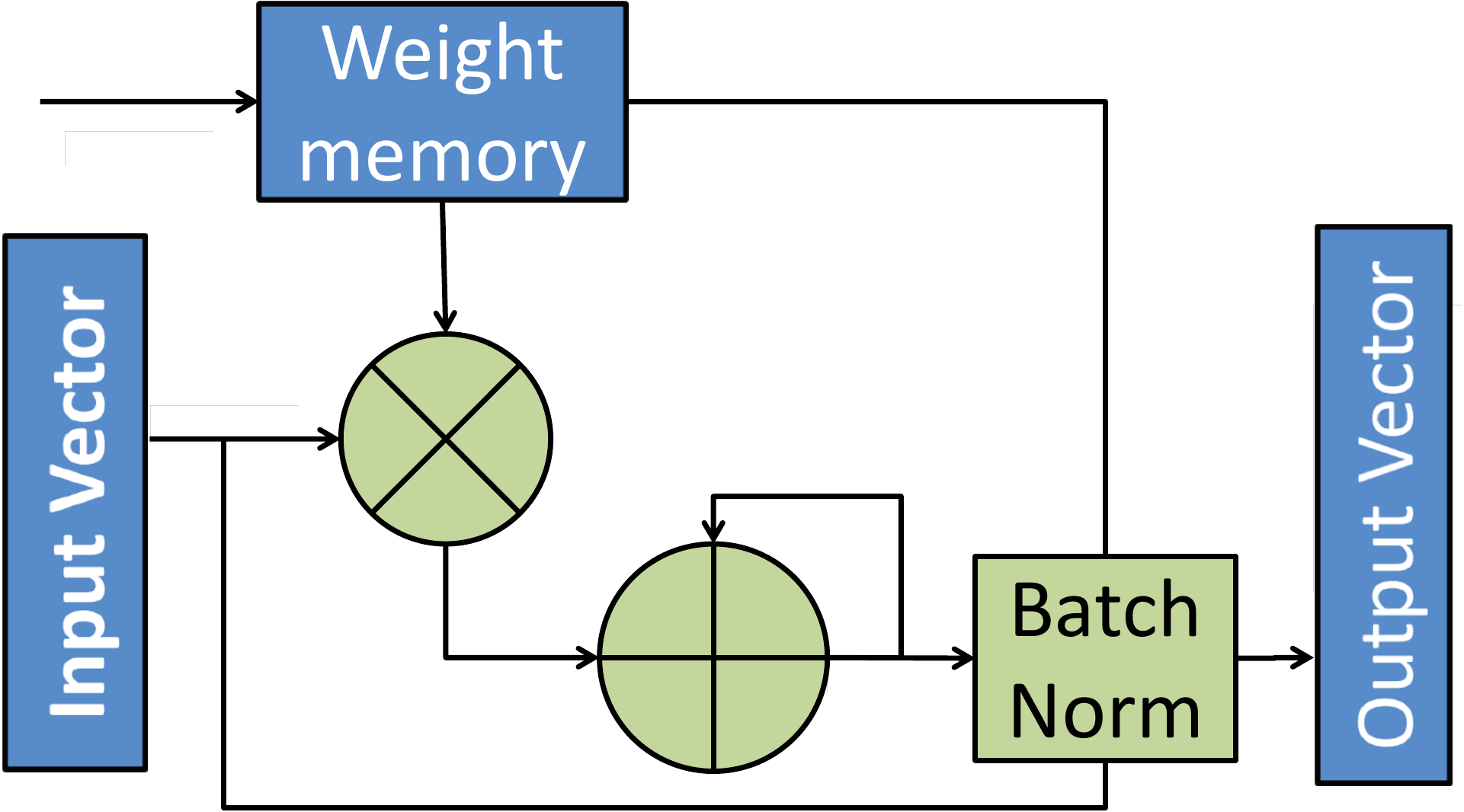}
		\captionof{figure}{Processing Element (PE)}
		\label{fig:fix-pe}
	\end{subfigure}
	\label{idontknowwhyitmustbehere}
	\begin{subfigure}[b]{0.45\linewidth}
		\centering
		\includegraphics[height=2.5cm]{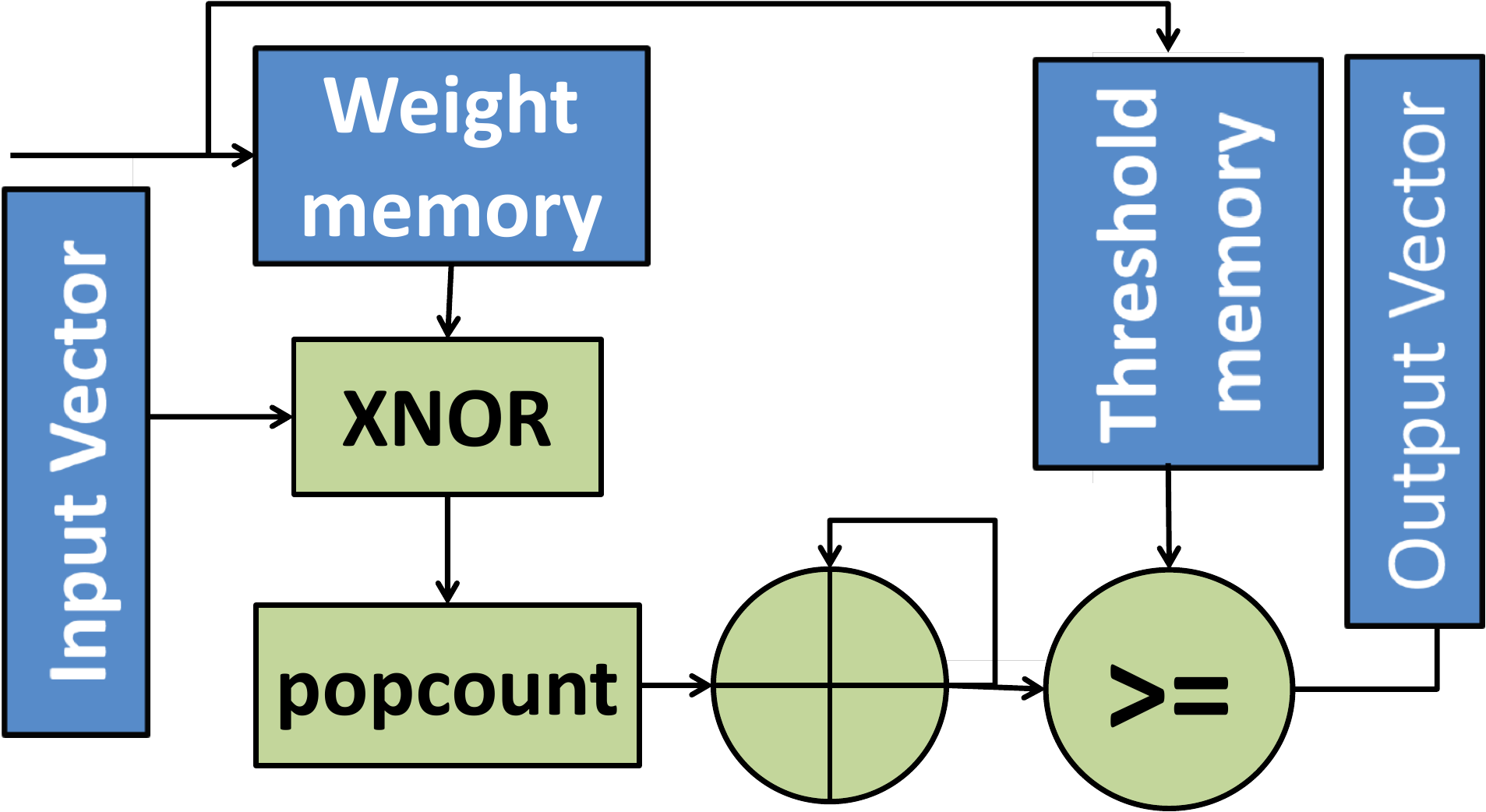}
		\captionof{figure}{Binarized Version PE}
		\label{fig:bin-pe}
	\end{subfigure}
	\caption{Hardware System Components for Neural Network Computation}
	\label{fig:finn-fig}
	\vspace{-20pt}
\end{figure*}

As shown in Figure \ref{fig:arch}, the overall system architecture used in this work is based on a data-flow framework for CNN inference called \FINN{} \cite{finn}. Network inputs are loaded from off-chip memory to layer-wise on-chip processing modules. After completing the feedforward computation, the classification outputs are finally transferred back to off-chip memory storage. As shown in Figure \ref{fig:arch}, each layer is mapped with an array of Matrix-Vector Operation Unit (MVOU) modules as shown in Figure \ref{fig:mvtu}.

Internally, the MVTU consists of an input and output buffer, and an array of \glspl{PE} each with a number of SIMD lanes. The number of PEs ($P$) and SIMD lanes ($S$) are configurable to control the throughout. A \gls{PE} can be thought of as a hardware neuron capable of processing $S$ synapses per clock cycle. Each PE receives exactly the same control signals and input vector data, but multiply-accumulates the input with a different part of the matrix.
Figure \ref{fig:fix-pe} shows the PE data-path for FIX/FP numbers and Figure \ref{fig:bin-pe} shows its counterpart for 1 bit binary numbers, which is used in \cite{finn}. Noticeably, the Multiplier-ACcumulate (MAC) structures (Figure \ref{fig:fix-pe}) for FIX/FP are replaced with XNOR and popcount structure for binary numbers. \textbf{Either MAC or XNOR/popcount is referred as ``operation" or ``fundamental operation" for its corresponding data type throughout this paper.} Please note, for higher precision parameters, the dataflow model is not necessarily feasible when the chips are not sufficiently large. These situations are beyond the assumption of this work and the related analysis is only for theoretical reference. 

\subsection{Hardware Cost Estimation Model}
\label{hw_cost_compute}

Based on the architecture described in last section, we propose our hardware cost estimation model for arbitrary parameter precision type and theoretically formulate the relationship between hardware cost and parameter precision type for the given architecture. Table~\ref{tab:op-cost} shows the average hardware cost per fundamental operation for each precision type. In order to get this table, different levels of parallelism for the MVOU has been tried in each data representation and we report an average value for fair comparisons. Because of the sharing of control logic, the average hardware cost for the basic operations can be different depending on the level of parallelism. So we mark the minimum and maximum cost of resources as ``min" and ``max" in the table and eventually use the average value of them for a more precise estimation. Look-Up-Tables (LUTs) and DSP blocks are both considered as hardware cost in this work. The average cost per operation, $C_{avg}$ is calculated as follows:
\begin{align}
\label{eq:avg_cost}
C_{avg} = \max(&\frac{LUTs/MAC}{LUT_{usage}*LUTs_{TOTAL}},\\
            &\frac{DSPs/MAC}{DSP_{usage}*DSPs_{TOTAL}})\ \ \mathrm{,}\nonumber
\end{align}
where $LUTs_{TOTAL}$ and $DSPs_{TOTAL}$ are separately the total available LUTs and DSPs on the target device and
$LUT_{usage}$ and $DSP_{usage}$ are separately the proportion of LUTs and DSPs that can be used for arithmetic on the target device. We've estimated $LUT_{usage}=0.7$ and $DSP_{usage}=1.0$ in this work.
$C_{avg}$ is the fraction of the target device resources that are used in average by a fundamental operation for each type and as such is a measure of scarcity of resource. Relative cost, $C_{rel}$, is used to compare the arithmetic cost of binarized networks against other precision types directly.
For example, if a Binary and an INT4 network have been trained to achieve the same level of accuracy, the INT4 network must have $5.38$ less operations to have the same accuracy / computation trade-off as the binarized one. \footnote{This assumes that both networks have the same memory footprint for their parameters.}

Interestingly, modelling computational cost this way means that INT16 has a \textit{lower} hardware cost than INT8, because it uses less LUTs/Op than INT8 and the proportion of DSPs that it uses per Op with respect to the total on the target device, a Xilinx Kintex UltraScale 115, is less than the proportion of LUTs/Op used by INT8. These resource usage data are calculated based on Vivado HLS 2016.3 synthesis reports. In this work, we only consider the default synthesis results from the compiler. Optimization to INT8 can be applied to recent Xilinx DSP blocks. This will improve INT8 performance but will not affect the correctness of our estimation model and hence not specially applied in our work. In essence, we assume a custom dataflow architecture generated for each specific network topology (different sizes for the compute arrays in different layers as shown in Fig.\ref{fig:arch}), meaning that the \dblquotes{one-size-fits-all} inefficiencies of loopback accelerators are avoided. As such, peak performance of a particular device is almost achievable in practice.

\subsection{Throughput Estimation Model}
\label{thpt_model}
Hardware cost is highly related to the system performance and computation efficiency. Theoretically, we formulate the relationship between inference throughput and hardware cost as follows:
\begin{equation}
    Throughput\approx \frac{Freq.}{\#OP\times C_{avg}+\Delta},
\label{thrpt}
\end{equation} where $Freq.$ is the working clock frequency, $\#OP$ is the number of operations required to compute a single NN input frame, which is a fixed value once network topology is determined. $\Delta$ stands for extra resource overhead used for control logic and $C_{avg}$ is defined in Eq.\ref{eq:avg_cost} as average hardware cost per operation. Because $C_{avg}$ in our estimation model is a ratio between required resource and the overall resource budget, $C_{avg}$ implies resource folding factor in order to get all computations done with available resource. We migrate and apply this folding effect to timing and interpret it as folding of clock cycles in unit time so that throughput can be estimated. As shown in Table \ref{tab:op-cost}, from binary values to 32-bit floating point values, the $C_{avg}$ is roughly getting higher due to the increasing hardware complexity except for the case where INT16 is more efficient than INT8 due to the explanation in Section \ref{hw_cost_compute}. Meanwhile, according to Eq.\ref{thrpt}, higher $C_{avg}$ brings down the throughput for the same network implementation on a specific device. This will be demonstrated in Section \ref{exp_res}.

According to our observation in real systems, as resource usage of the target device increase, the models become more accurate. For concrete examples, we compare results from our estimation model to real implementation from Fraser ~et~al~\cite{fraser2017scaling_s}. The measured GOps/s for their cnn(1/2) and cnn(1) models are 1856 and 7407. According to our estimation model, the estimated minimum performance for the corresponding models are 2051 and 8596, which are 35\% and 16\% difference. The discrepancy between estimated and measured performance could be due to the following factors: 1. Difference in clock frequency between estimated and measured models. 2. An underestimation of the control logic overhead when a small portion of the target device is used. 3. the model doesn't take into account that the first layer has 8-bit pixel images as inputs.

\section{Experimental Evaluation}
\label{experiments}



We tested on 6 precision types: 1-bit binary values (Binary), fixed point representations with 2-bit (INT2), 4-bit (INT4), 8-bit (INT8), 16-bit (INT16) and single-precision floating point values (FP32). 2 bits are reserved for the integer part and rest for fractional part ($FL$). The fully-connected NNs are tested on the MNIST dataset. CNNs are tested on CIFAR-10 \cite{cifar10} and ImageNet \cite{krizhevsky2012imagenet} datasets. All input images are expressed in 8-bit fixed point numbers.

We used Fully-Connected(FC) and CNN models in our experiments. FC is a reference network topology with 3 hidden layers with each containing 4096 hidden neurons fully connected to its proceeding layer. For \gls{CNN}, the reference topology is the VGG-16 inspired model \cite{vgg16}, which contains a succession of (3$\times$3 convolution, 3$\times$3 convolution, 2$\times$2 maxpool) layers repeated three times with 128-256-512 channels, followed by two fully-connected layers with 1024 neurons in each. For ImageNet tasks, we use AlexNet \cite{krizhevsky2012imagenet} as baseline model. In terms of activation function, all the precision options are trained with \gls{RELU} and \gls{HARDTANH} and the best accuracy results are used to report the performance. Additionally, 5 values for scaling factor $s$ are applied to the reference networks in order to expand or shrink the reference topology in a specific ratio. The values are 0.03125, 0.0625, 0.125, 0.25, 0.5, 1. For example, all tested FC networks have the same number of hidden layers, but with $1024*s$ neurons correspondingly. Similarly to CNNs, scaling factors are multiplied with the number of filters in each conv layer, but they do not change the depth of the topology. For ImageNet tasks, smaller models provide unacceptably low accuracy, so we only report the results of 0.25, 0.5 and 1.

In this work, we use Xilinx Kintex UltraScale 115 as the target FPGA device. The working clock frequency is 250 MHz. In terms of metrics, \textbf{throughput} is measured in this work as frames per second and a frame is an image fed to a neural network. \textbf{Hardware Cost} is studied through computational resources and block ram (BRAM) usage. Since we are not competing for the best model accuracy, better classification results can be achieved if using other optimization techniques, which can be orthogonal to the training strategy used in this work. To make fair comparison, we train all experiments for each dataset/topology with the same hyper-parameters including number of epochs, learning rate decay strategy etc.

\subsection{Experimental Results}
\label{exp_res}

\begin{figure}
\vspace{-10pt}
\centering
\includegraphics[height=3in]{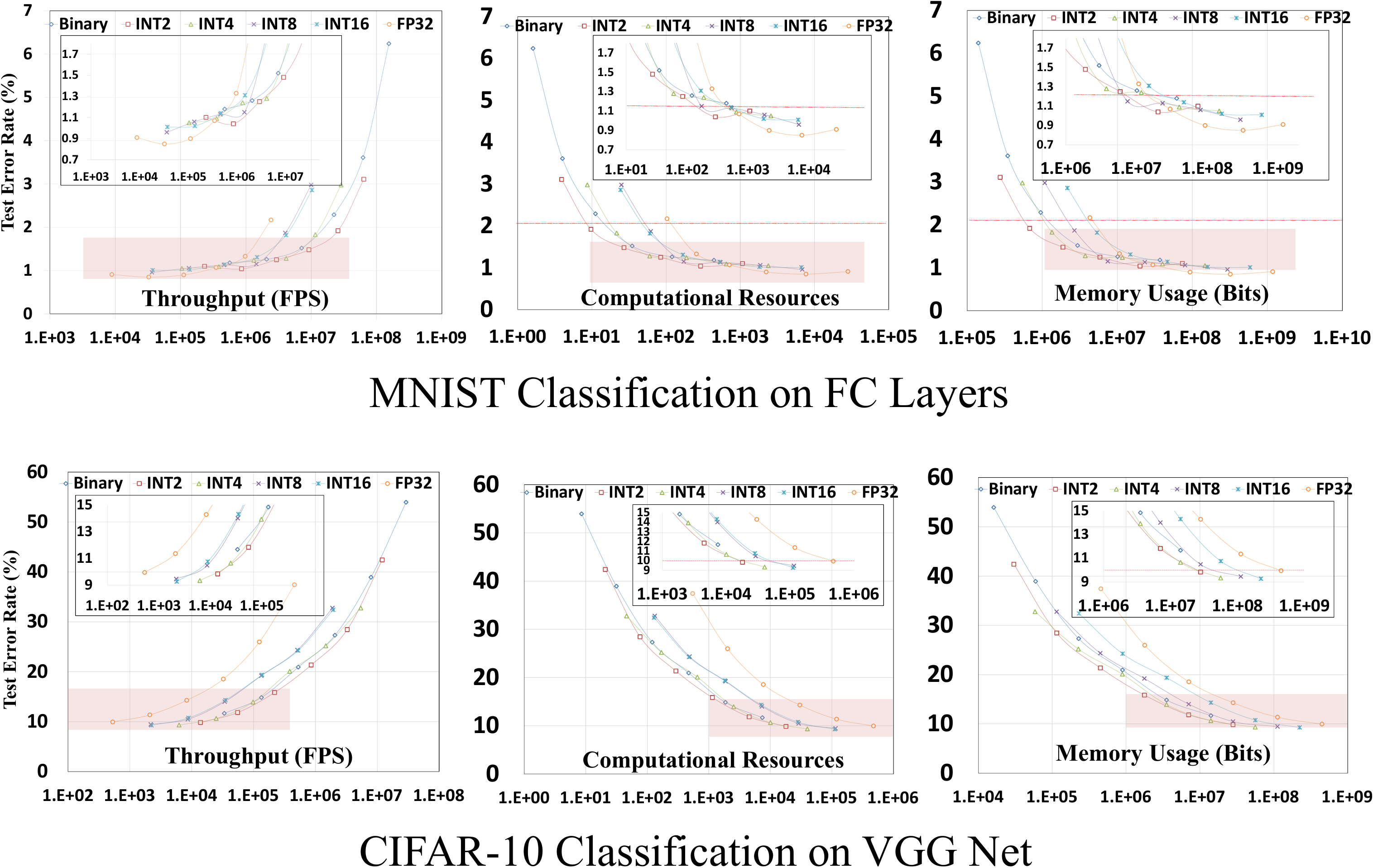}\\
\caption{Experimental Results for CIFAR10 and MNIST Classification}
\vspace{-20pt}
\label{cifar10-mnist}
\end{figure}

\begin{figure}
\begin{centering}
\includegraphics[height=3in]{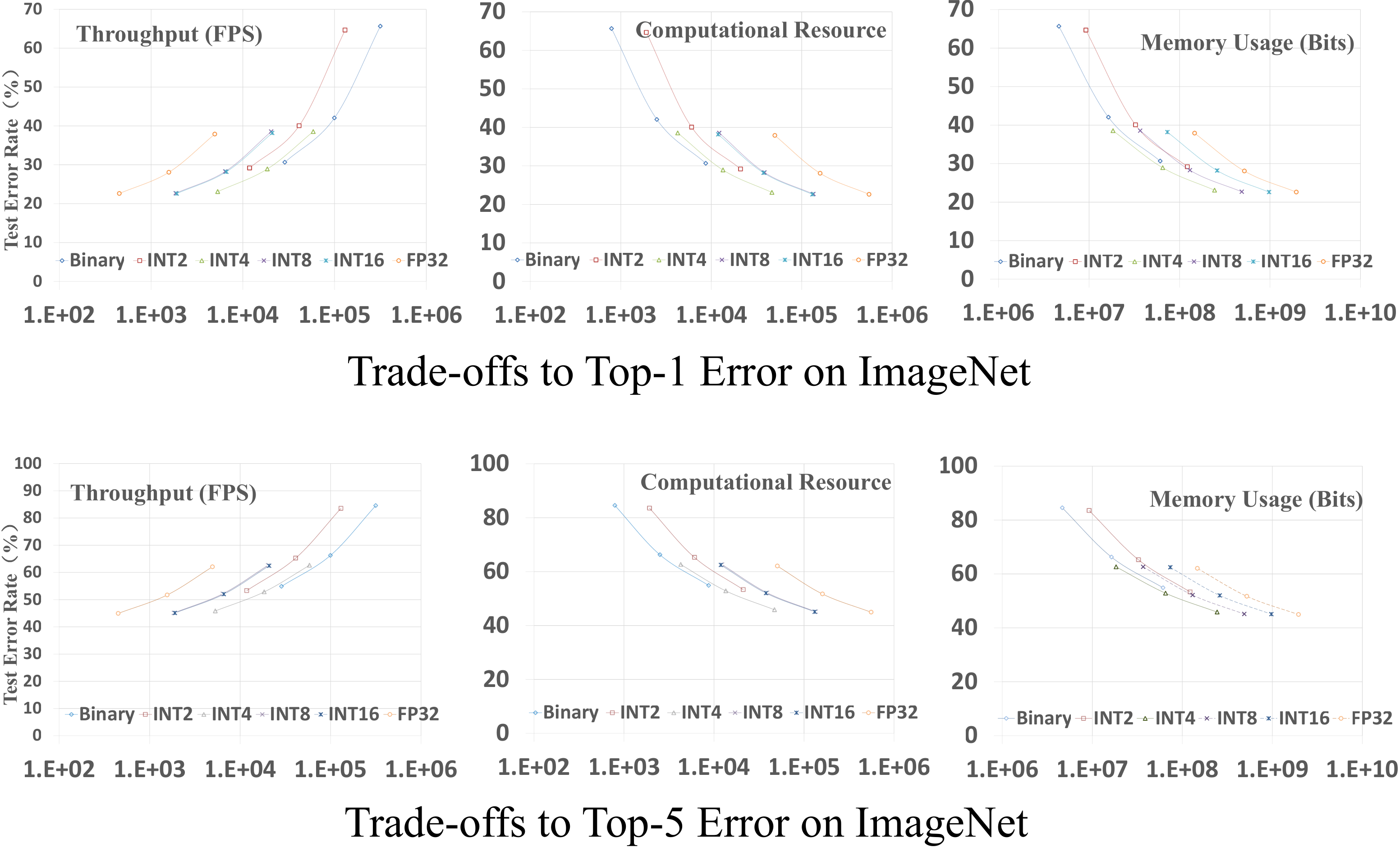}\\
\caption{Experimental Results for ImageNet Classification on AlexNet}
\label{imagenet}
\end{centering}
\vspace{-40pt}

\end{figure}

The results shown in this section are based on our estimation models. Figure \ref{cifar10-mnist} and \ref{imagenet} show the trade-off curves for different dataset and network combinations. Each curve indicates a data representation for both weight and activation. Each marker on the curve shows the result for a network with a specific scaling factor. In Figure \ref{cifar10-mnist}, the areas highlighted in red colour are emphasized in an attached zoom-in view in order to show more information about regions for high classification accuracy regions, which may deliver more insights that global trends cannot display.

\textbf{MNIST on FC Layers} From Figure \ref{cifar10-mnist}, we can see that FP32 delivers the highest accuracy in 3 of its topology options and Binary provides best options in terms of hardware efficiency with a much higher accuracy drop (6.24\%) compared to the best FP32 results. In general, INT2, FP32 and INT4 dominate the Pareto Frontier.
From the zoom-in views, a noticeable observation is that among solutions that give no higher than 1.2\% classification error, which is the best achievable result for binary, INT2, INT4 and even FP32 can all provide more efficient solutions than Binary in terms of memory usage. The reason for this is that Binary requires a larger topology and more computation to achieve the same model accuracy. For example, only $4096*0.125=512$ neurons are needed in each hidden layer for FP32 to achieve 1.02\% error while 4096 neurons per hidden layer are needed for Binary to achieve a similar error of 1.2\%. Required memory for Binary is 37.0 Mb and 29.8 Mb for FP32. 

Noticeably, with a relatively small budget of BRAM smaller than 1Mb, only Binary, INT2 and INT4 are feasible options for hardware implementations while INT2 can achieve the highest accuracy (98.1\%). If comparing the representability of Binary and FP32 by looking at the solutions with best accuracy for each, Binary requires only 27 Mb memory for an accuracy of 98.8\% while FP32 needs 1.2 GB for only a 0.25\% higher accuracy. Additionally, if setting the accuracy goal as 98\% on MNIST task (red dotted line on the global figure), INT2 provides the most efficient option in terms of computational resource and memory usage. Meanwhile, Binary at least requires 6.2$\times$ more computational resources and 7.8$\times$ more memory compared to the optimal INT2 option. Besides, INT4 also provides more resource/memory efficient options than Binary.

\begin{table*}[]
\vspace{-25pt}
\centering
\caption{Memory Saving and Model Accuracy Comparison Between Our Work and RER\cite{rer}, LRD\cite{denil1014}, DK\cite{dk}, HashNet\cite{chen15}, Q-CNN\cite{wu0516} and Q-CNN(EC)\cite{wu0516} on MNIST Classification Task with 784-1000-1000-1000-10 FC Networks}
\label{ext_compare}
\begin{tabular}{|c|c|c|c|c|c|c|c|c|c|c|c|}
\hline
                   & \multicolumn{5}{c|}{Ours}                                                                                                                                      &                       &                       &                      &                           &                         &                             \\ \cline{2-6}
\multirow{-2}{*}{} & Binary                                     & INT2                & INT4          & INT8                                 & INT16                                & \multirow{-2}{*}{RER} & \multirow{-2}{*}{LRD} & \multirow{-2}{*}{DK} & \multirow{-2}{*}{HashNet} & \multirow{-2}{*}{Q-CNN} & \multirow{-2}{*}{Q-CNN(EC)} \\ \hline
Mem. Saving        & {\color[HTML]{FE0000} \textbf{32$\times$}} & \textbf{16$\times$} & 8$\times$     & 4$\times$                            & 2$\times$                            & 8$\times$             & 8$\times$             & 8$\times$            & 8$\times$                 & 12.1$\times$            & 12.1$\times$                \\ \hline
Error           & 1.5                                        & 1.25                & \textbf{1.16} & {\color[HTML]{FE0000} \textbf{1.13}} & {\color[HTML]{FE0000} \textbf{1.13}} & 1.24                  & 1.77                  & 1.26                 & 1.22                      & 1.34                    & 1.19                        \\ \hline
\end{tabular}
\vspace{-20pt}
\end{table*}
Moreover, our low-precision training strategy allows memory saving when conducting inference, which achieves a very similar effect of network compression. We compare our results with several state-of-the-art compression works on the same dataset (MNIST) and the same network topology. Table \ref{ext_compare} shows that without using any compression techniques, our INT4, INT8, INT16 results can achieve higher model accuracy than the other methods \footnote{The reason that the particular 784-1000$\times$3-10 structure is selected in Table \ref{ext_compare} is that it is the only structure that is reported in all mentioned works. We compare different methods on the same structure in the same classification task for fair comparisons on memory and accuracy.}. Meanwhile, INT2 and Binary achieve higher memory saving rate and still keep competitive accuracy. As highlighted in red colour, our results either achieve best compression rate or highest accuracy on the exactly same network topology compared to the other state-of-art results.

\textbf{CIFAR10 on VGGNet} Second row in Figure \ref{cifar10-mnist} shows trade-offs for CIFAR10 classification with VGGNets. Noticeably, INT4, INT8 and INT16 provide very close best accuracy and all higher then FP32 alternative. The rounding noise introduced in parameter quantization may help to improve the classification accuracy in this particular case. Similarly, Binary provides the most efficient solution among all precision types but with much higher error (54\%). INT2 and INT4 provide high accuracy options with relatively higher throughput, lower resource and memory usage, as shown in the zoom-in views. They are considered as optimal parameter data type as they contribute most of the Pareto-efficient options. FP32 options are not advantageous on either model accuracy or hardware cost because of the high complexity. As shown by the red dotted lines in the zoom-in views, for the range where accuracy is higher than 90\%, it is INT2, rather than Binary, that provides the most efficient options in terms of computational resource and memory usage. Specifically, for 91\% accuracy, INT2 provide 13.9K FPS which is 26$\times$ higher than FP32, 6 $\times$ higher than INT8 and INT16 alternatives. On the other hand, Binary requires a larger topology to achieve the same level of accuracy compared to the other alternatives by 1 scaling factor. But it presents only 1.7\% accuracy degradation with 32$\times$ less memory and 63.8$\times$ less computational resource requirement if we compare the most accurate options provided by Binary and FP32.

\textbf{ImageNet on AlexNet} ImageNet tasks show clearer relative positions among curves in Fig. \ref{imagenet}. This can be caused by the higher complexity in the classification tasks compared to MNIST and CIFAR10. Noticeably, Binary and INT2 solutions cannot achieve comparable model accuracy to other data types as they are in MNIST and CIFAR10 tasks. As shown in all figures, there is an accuracy gap between these two types and the others. Some very recent works like \cite{hwgq} try to target on this accuracy gap by optimizing the quantization function for parameters with extremely low bitwidth. This topic is very interesting and definitely deserves more efforts, but it is beyond the scope of this paper. In particular, INT4, compared to FP32, provide solution with 8$\times$ memory saving and 11.8$\times$ higher throughput with only less than 1\% accuracy drop. Similarly, INT8 can provide a 4.3$\times$ higher throughput solution with no accuracy loss compared to FP32. Therefore, FP32 again loses its advantage on either model accuracy or hardware efficiency. In general, INT4, INT8 and INT16 and FP32 present accuracy with negligible difference. However, INT4 has the best trade-off due its less memory and computational resource requirements as well as higher system throughput possibly provided. 

\section{Summary and Conclusion}
\label{conclusion}
In this work, we firstly introduce our quantization training strategy that allows training NNs with arbitrary parameter precision. Then, we propose our hardware cost and throughput estimation models. Finally, we conduct our experiments in the exploration space consist of 6 different data types, 3 different NN models and 3 different benchmarks. We found that Binary does not necessarily provide hardware solutions with highest efficiency due to larger amount of parameters required for Binary to achieve the same level of model accuracy with its high precision alternatives. Within our studied cases, INT2 and INT4 generally provide better trade-offs in small image classification tasks, MNIST and CIFAR10, while INT4 provide the best trade-offs among all other types in ImageNet tasks. More insightful observations have been pointed out in Section \ref{experiments}, which hopefully can be helpful to reduced-precision NN system design on reconfigurable hardware.


\section*{Acknowledgments}

The authors from Imperial College London would like to acknowledge the support of UK’s research council (RCUK) with the following grants: EP/K034448, P010040 and N031768. The authors from The University of Sydney acknowledge support from the Australian Research Council Linkage Project LP130101034.

\bibliography{references}
\bibliographystyle{ieeetr}
\end{document}